\documentclass[11pt]{article}

\usepackage[utf8]{inputenc}
\usepackage[T1]{fontenc}
\usepackage[margin=1in]{geometry} 
\usepackage{authblk} 

\usepackage{booktabs}
\usepackage{multirow}
\usepackage{graphicx}
\usepackage{amssymb}
\usepackage{wrapfig}
\usepackage{bbm}
\usepackage[accsupp]{axessibility} 

\usepackage[pagebackref,breaklinks,colorlinks,citecolor=blue,linkcolor=red]{hyperref}
\usepackage{orcidlink}

\newcommand{\modelname}{GenOpticalFlow}

\title{\textbf{\modelname: A Generative Approach to Unsupervised Optical Flow Learning}}

\author[1]{Yixuan Luo\thanks{Equal contribution}}
\author[2]{Feng Qiao\protect\footnotemark[1]}
\author[2]{Zhexiao Xiong}
\author[1]{Yanjing Li}
\author[2]{Nathan Jacobs}

\affil[1]{University of Chicago}
\affil[2]{Washington University in St. Louis}

\date{} 
\DeclareUnicodeCharacter{202F}{\,}
\begin{document}

\maketitle

\begin{abstract}
Optical flow estimation is a fundamental problem in computer vision, yet the reliance on expensive ground-truth annotations limits the scalability of supervised approaches. Although unsupervised and semi-supervised methods alleviate this issue, they often suffer from unreliable supervision signals based on brightness constancy and smoothness assumptions, leading to inaccurate motion estimation in complex real-world scenarios. To overcome these limitations, we introduce \textbf{\modelname}, a novel framework that synthesizes large-scale, perfectly aligned frame--flow data pairs for supervised optical flow training without human annotations. Specifically, our method leverages a pre-trained depth estimation network to generate pseudo optical flows, which serve as conditioning inputs for a next-frame generation model trained to produce high-fidelity, pixel-aligned subsequent frames. This process enables the creation of abundant, high-quality synthetic data with precise motion correspondence. Furthermore, we propose an \textit{inconsistent pixel filtering} strategy that identifies and removes unreliable pixels in generated frames, effectively enhancing fine-tuning performance on real-world datasets. Extensive experiments on KITTI2012, KITTI2015, and Sintel demonstrate that \textbf{\modelname} achieves competitive or superior results compared to existing unsupervised and semi-supervised approaches, highlighting its potential as a scalable and annotation-free solution for optical flow learning. We will release our code upon acceptance.

\end{abstract}

\section{Introduction}
\label{sec:intro}

\begin{figure}[t]
    \centering
    \includegraphics[width=\columnwidth]{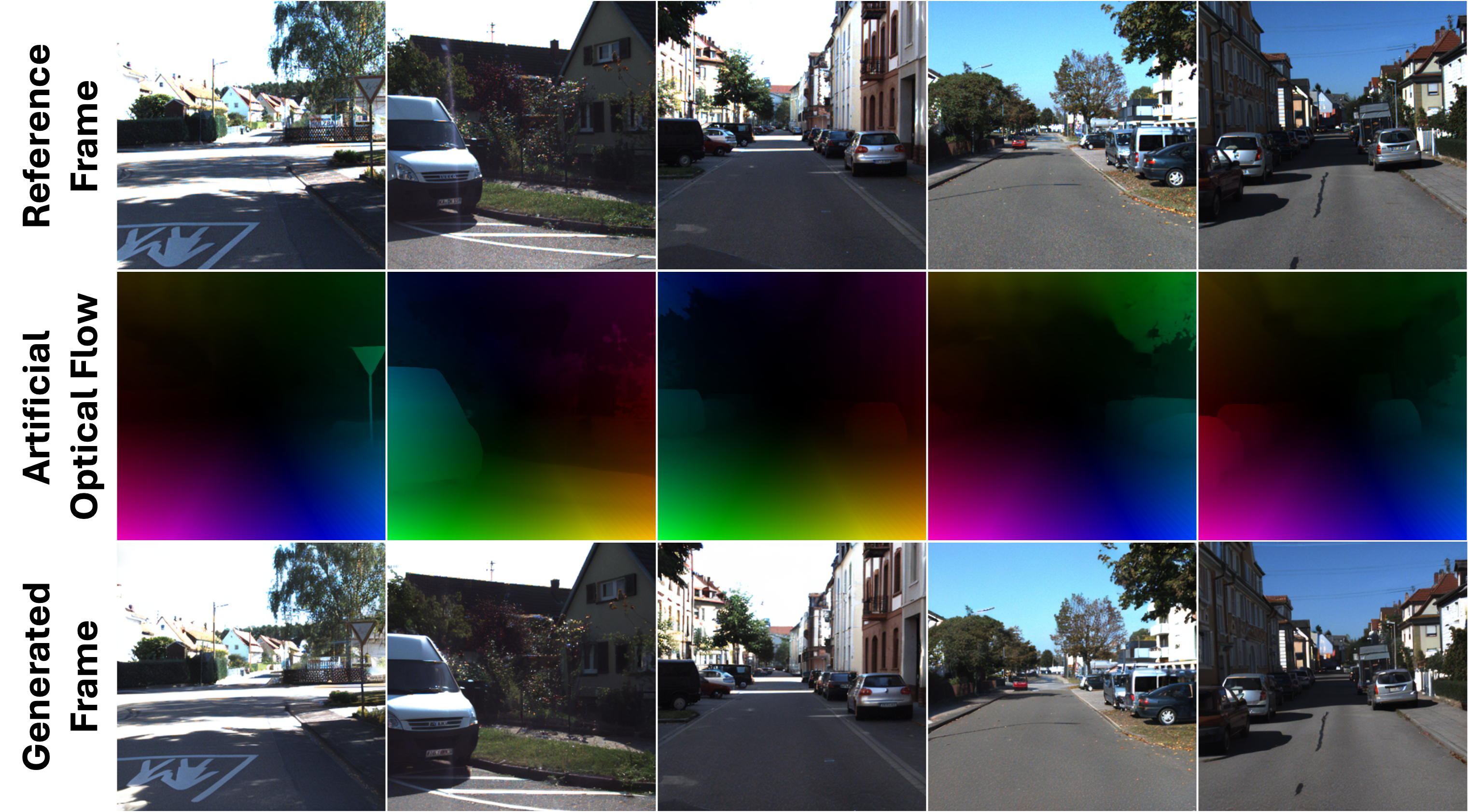}
    \caption{Visualization of the synthesized data triplet on KITTI2012, including the reference frame \( I_{t} \), the artificially generated optical flow \( \tilde{F}_{t \rightarrow t+1} \), and the conditioned next-frame generation prediction result \( \tilde{I}_{t+1} \).
    }
    \label{fig:generation-sample}
\end{figure}

Optical flow estimation is a fundamental task in computer vision that aims to capture the 2D pixel-level motion between consecutive video frames. This task plays a crucial role in numerous downstream applications, including autonomous driving~\cite{geiger2013vision,menze2015object,janai2020computer}, frame interpolation~\cite{liu2020video,huang2020rife,xu2019quadratic}, action recognition~\cite{sun2018optical,piergiovanni2019representation}, and video understanding~\cite{ye2022deformable}.

With the rapid advancement of deep learning, optical flow estimation has gradually shifted from traditional hand-crafted optimization over dense displacement fields between image pairs~\cite{horn1981determining,lucas1981iterative,brox2004high,zach2007duality} to supervised learning with neural networks trained on ground-truth optical flows~\cite{dosovitskiy2015flownet,ilg2017flownet,sun2018pwc,hui20liteflownet2,teed2020raft}. Although supervised learning methods have demonstrated superior performance, obtaining accurate ground-truth optical flow labels in real-world scenarios remains challenging. Obtaining such data typically requires manual calibration, which is both time-consuming and expensive~\cite{han2022realflow,yuan2022optical,im2022semi}. This fundamental bottleneck has limited the large-scale deployment of supervised optical flow estimation methods.

To address this key limitation, recent approaches~\cite{meister2017unflowunsupervisedlearningoptical,liu2019selflowselfsupervisedlearningoptical,liu2019ddflowlearningopticalflow,wang2018occlusion,luo2021upflow} have exploited unsupervised learning, thereby avoiding the substantial requirement for annotated ground-truth optical flow. Rather than relying on explicit labels, unsupervised optical flow methods typically leverage two inherent properties of consecutive frames: brightness constancy and spatial smoothness. These principles form the basis for designing appropriate loss functions that guide the learning process. Consequently, even without ground-truth flow for training, unsupervised techniques have demonstrated outstanding performance and generalization capacity.

Despite their advantages in reducing annotation costs and improving scalability, unsupervised optical flow methods still lag behind supervised approaches. Their key limitation lies in the indirect supervision signal: instead of learning from ground-truth motion fields, they rely on photometric reconstruction losses based on brightness constancy and spatial smoothness assumptions~\cite{jonschkowski2020matters,wang2018occlusion,meister2017unflowunsupervisedlearningoptical,huang2023self}. These assumptions, however, often fail under real-world conditions with illumination changes, motion blur, or occlusions, making the loss a noisy proxy for true motion. Occluded regions further lack valid correspondence, leading to oversmoothed or ambiguous flow predictions~\cite{janai2018unsupervised,liu2019selflowselfsupervisedlearningoptical,liu2019ddflowlearningopticalflow}. Moreover, since the objective is purely appearance-based, unsupervised models struggle to capture higher-level semantic motion patterns that supervised methods can implicitly learn. Consequently, despite notable progress, a consistent performance gap remains due to the inherent unreliability of unsupervised supervision signals~\cite{wang2018occlusion}.

To bridge this gap, recent work~\cite{feng2023flowdaunsuperviseddomainadaptive,lai2017semi,yan2020optical,zhang2022clip} explores semi-supervised and domain-adaptive strategies that combine limited ground-truth supervision with unlabeled data. These hybrid strategies show that even minimal reliable supervision can greatly narrow the gap, yet they still rely on labeled data and thus fail to eliminate dependence on manual annotations and calibration.

An emerging paradigm aims to overcome the limitations of unreliable proxies (e.g., photometric loss) and costly manual annotations by harnessing vision generation models~\cite{ho2020ddpm,song2020denoising,rombach2022high,dhariwal2021diffusion,ho2022classifier}. These models, particularly diffusion-based approaches, excel at synthesizing complex, high-fidelity images from a noise prior. Their power lies in their ability to be guided by conditioning information, such as text prompts~\cite{rombach2022high} or class labels~\cite{ho2022classifier}. This mechanism allows for precise control over the generative process, enabling the creation of diverse and photorealistic images that align with specific semantic inputs.

Motivated by this, we proposed \textbf{\modelname}, shifting the focus from inferring flow from unlabeled real-world videos to synthesizing large-scale frame-and-flow data pairs for enhancing model performance with supervised training. The key advantage of our approach lies in its ability to generate perfect, pixel-level ground truth by design: a generative model, conditioned on an existing frame and an explicit optical flow field, synthesizes a corresponding, perfectly aligned subsequent frame. In real-world domains, however, we only have frames, lacking the accurate ground truth optical flow needed for this conditioning. To generate these conditioning flows, we adapt architectures originally designed for depth estimation~\cite{yang2024depth,bhat2023zoedepth,ricci2018monocular,zhao2022monovit}, repurposing them to produce plausible temporal pixel shifts instead of spatial disparities. Given a source frame and this synthetic optical flow, our framework can then generate the accurately aligned next frame, completing the synthetic data sample. This synthesis process creates a virtually infinite supply of accurately labeled data at minimal cost, sidestepping real-world challenges like occlusions and illumination variations. By eliminating the need for manual annotations, it provides a clean, reliable, and scalable signal for training robust optical flow networks.

In summary, our contributions are as follows.
\begin{itemize}
    \item We propose \textit{\modelname}, a framework that leverages a depth estimation model and a vision generation model to synthesize labeled optical flow data pairs, enabling supervised training from unlabeled videos.
    \item We propose a novel next-frame prediction architecture conditioned on a given frame and optical flow, capable of achieving accurate pixel-level motion prediction. Compared to other conditioning frame generation methods, \modelname achieves SOTA performance on the generation quality. 
    \item Within \textit{\modelname}, we propose a novel unsupervised diagram for the optical flow task, which leverages the synthetic data triplet and fine-tunes the model in a supervised manner. Across diverse methods, \modelname achieves consistent improvements across seven different frameworks, reducing EPE by 1.49 and Fl-all by 7.00, respectively.
\end{itemize}
\section{Related Works}
\label{sec:relatedworks}

\paragraph{Optical Flow Models}
Optical flow estimation, which aims to recover dense pixel-wise motion fields between consecutive frames, has evolved from classical variational formulations such as Horn–Schunck and Lucas–Kanade~\cite{horn1981determining,lucas1981iterative}, built upon brightness constancy and smoothness assumptions, to data-driven supervised models that learn motion representations directly from annotated datasets. Early deep networks like FlowNet and FlowNet2~\cite{dosovitskiy2015flownet,ilg2017flownet} demonstrated the feasibility of end-to-end optical flow learning, while PWC-Net~\cite{sun2018pwc} integrated pyramid processing, warping, and cost volumes to achieve higher accuracy and efficiency. Later architectures, notably RAFT~\cite{teed2020raft} and WAFT~\cite{wang2025waftwarpingalonefieldtransforms}, advanced the field with dense all-pairs correlations and iterative refinement, setting new performance benchmarks. More recent supervised models such as GMA~\cite{jiang2021gma}, GMFlow~\cite{xu2022gmflow}, and FlowFormer~\cite{huang2022flowformer} further enhance global correspondence modeling and long-range context aggregation, achieving state-of-the-art results across standard benchmarks. Most recently, WAFT~\cite{wang2025waftwarpingalonefieldtransforms} introduces warping-alone field transforms tailored for optical flow.

\paragraph{Unsupervised Optical Flow Learning}
Unsupervised optical flow estimation has matured as a compelling alternative to supervised methods by leveraging photometric reconstruction, forward–backward consistency, temporal continuity and semantic or geometric priors rather than dense ground‑truth flow. Early multi‑frame formulations such as Janai et al.~\cite{janai2018unsupervised} introduced occlusion‑aware and multi‑frame warping losses. The sequence‑aware self‑teaching strategy of SMURF~\cite{stone2021smurf} further improved accuracy by adapting the architecture of RAFT for unsupervised settings. Regularization techniques such as the teacher–student content‑aware regularizer in “Regularization for Unsupervised Learning of Optical Flow”~\cite{wang2023regularization} enhanced cross‑dataset generalization. StereoFlowGAN~\cite{Xiong_2023_BMVC} co-trains stereo and flow with unsupervised domain adaptation to better transfer from synthetic to real data. More recent methods inject higher‑level cues: for instance, SemARFlow~\cite{yuan2023semarflow} uses semantic segmentation masks to refine boundaries in autonomous driving scenes, and UnSAMFlow~\cite{li2024unsamflow} integrates object‑level masks from the Segment Anything Model (SAM) to sharpen motion boundaries. Simultaneously, spatial‑temporal dual‑recurrent modeling for dynamic environments was proposed in Sun et al.~\cite{sun2024temporal}, which handles occlusion and content variation via temporal priors. These advances show that unsupervised approaches are closing the gap to supervised models while allowing large‑scale training without manual annotations.

\paragraph{Conditioned Image Generation}
Recent advances in image generation have been driven by denoising diffusion probabilistic models (DDPMs)~\cite{ho2020ddpm} and latent diffusion frameworks (LDMs)~\cite{rombach2022high}, which refine the generative process via iterative denoising of a Gaussian noise sequence. Early DDPM work demonstrated that generative modelling via a forward-noise and reverse-denoising chain could match or surpass GANs in image quality~\cite{song2020score,ho2020ddpm}. To accelerate sampling and improve interpolation, non-Markovian variants such as DDIM were proposed~\cite{song2020denoising}. However, pixel-space diffusion remains computationally expensive. The seminal LDM work by Rombach et al.~\cite{rombach2022highresolution} alleviates this by applying the diffusion process in a learned latent space: an autoencoder encodes images into a low-dimensional latent, the diffusion U-Net operates in that space, and finally a decoder reconstructs the image; this approach reduces compute while preserving fidelity. 

Building on these foundations, coordinate-conditioned generative models such as GenWarp~\cite{seo2024genwarp} and GenStereo~\cite{qiao2025genstereo} condition the generator on continuous spatial coordinates to produce high-quality, view-consistent images. While GenWarp achieves strong semantic preservation for single-image novel view synthesis, its warping-based formulation prioritizes perceptual consistency rather than enforcing pixel-level motion accuracy. GenStereo, on the other hand, is tailored to horizontally aligned stereo pairs and primarily models disparity-induced viewpoint changes under constrained camera geometry.

In contrast, \modelname{} is explicitly designed for optical flow supervision rather than view-consistent image synthesis. Our framework generates geometrically grounded and pixel-aligned motion fields that serve as pseudo ground-truth flow, enabling downstream flow refinement. Unlike stereo-focused methods that assume epipolar constraints or limited baseline motion, \modelname{} handles general 2D motion induced by arbitrary camera translations, producing dense correspondence fields beyond disparity-only settings. This shift from appearance-conditioned rendering to geometry-consistent motion supervision fundamentally differentiates our approach from prior coordinate-conditioned generative models.

\section{Methods}

\subsection{Problem Formulation}

\paragraph{Optical Flow Estimation}
Given two consecutive RGB frames \( I_{t}, I_{t+1} \in \mathbb{R}^{H\times W\times 3} \), optical flow estimation aims to compute the dense optical motion field \( F_{t\rightarrow t+1} \in \mathbb{R}^{H\times W\times 2} \) that represents the per-pixel displacement from \( I_{t} \) to \( I_{t+1} \). Depending on whether ground-truth flow annotations are available, the task can be categorized as supervised or unsupervised optical flow estimation.

\paragraph{Next-Frame Generation Conditioned on Optical Flow}
Given an RGB frame \( I_{t} \in \mathbb{R}^{H\times W\times 3} \), an optical flow field \( F_{t\rightarrow t+1} \in \mathbb{R}^{H\times W\times 2} \), and a conditional image generation model \( \mathcal{G} \), the next-frame generation task aims to synthesize the subsequent frame \( \tilde{I}_{t+1} = \mathcal{G}(I_t, F_{t\rightarrow t+1}) \) by leveraging both the appearance and motion information from the given inputs. The objective is to minimize the pixel-level discrepancy between the generated frame \( \tilde{I}_{t+1} \) and the ground-truth frame \( I_{t+1} \).

\paragraph{Geometric Optical Flow Synthesis}
Given an estimated depth map \( D_{t} \in \mathbb{R}^{H \times W} \) and a set of camera parameters \( P \), the goal is to obtain accurately generated optical flow for conditioning the synthesis of the next frame. To achieve this, we utilize a novel view synthesis model \( S \) that can produce reliable synthetic optical flow through forward warping, denoted as \( \tilde{F}_{t \rightarrow t+1} = S(D_t, P) \).

\subsection{Conditioned Frame Generation}
\label{Sec.Methods.ConditionedFrameGeneration}
\begin{figure*}[t]
    \centering
    \includegraphics[width=\textwidth]{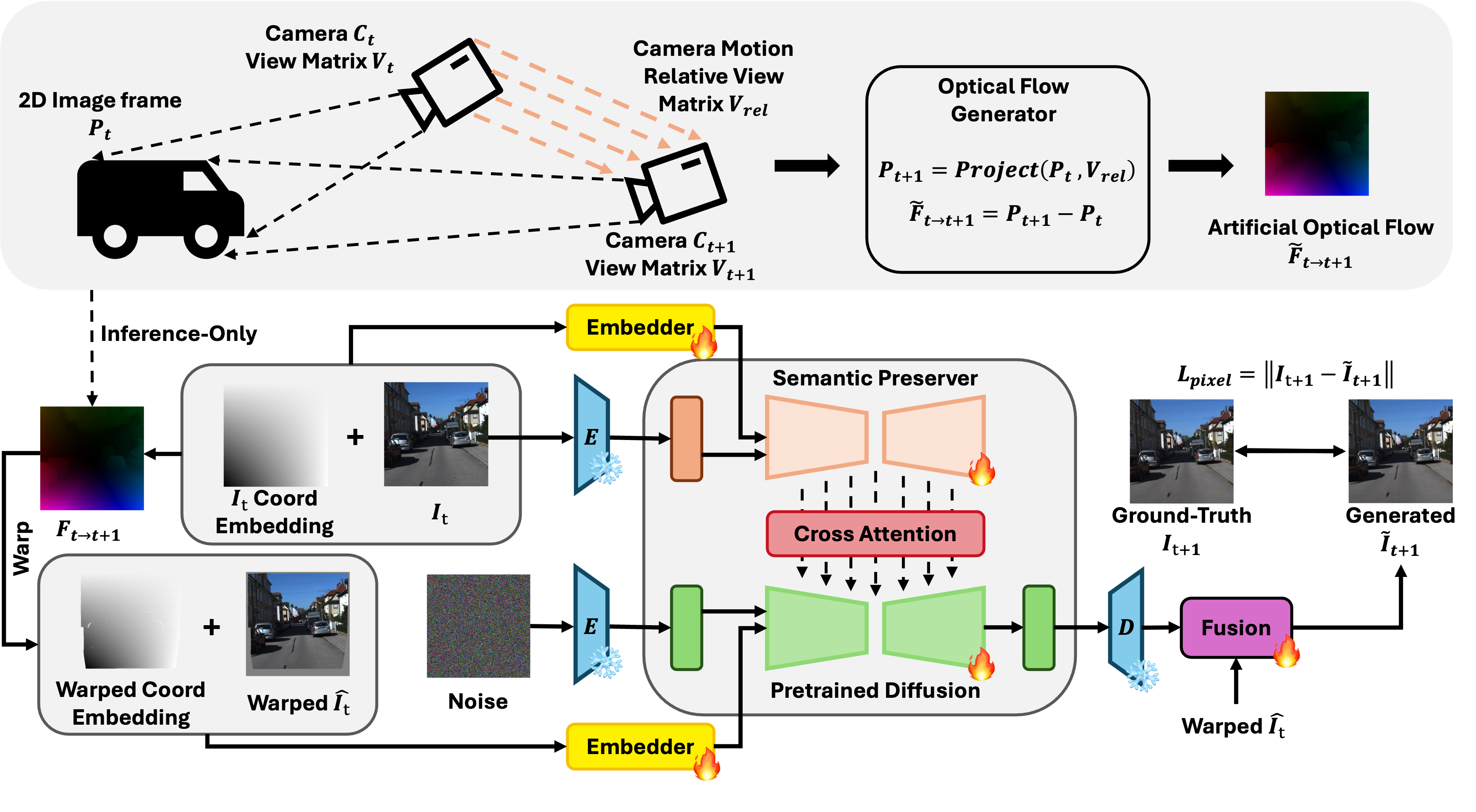}
    \caption{Overview of the conditioned next-frame framework and artificial optical flow generation. Given an input view and its corresponding optical flow, our framework constructs two types of embeddings: a 2D coordinate embedding of the input view and a warped coordinate embedding of the target view derived from the optical flow. A semantic preserver network extracts high-level semantic features from the input view, while a diffusion model conditioned on these embeddings learns the geometric warping necessary to generate the novel view and accurately align pixel-level motion. To further enhance spatial correspondence, we augment the self-attention mechanism with cross-view attention and jointly aggregate features across views. Notably, ground-truth optical flow is used only during pre-training, while synthetic datasets are constructed using artificial optical flow produced via a geometry-based camera-motion model.}
    \label{fig:conditioned-frame-generation}
\end{figure*}

Given an RGB frame \( I_{t} \in \mathbb{R}^{H \times W \times 3} \) and a corresponding guiding optical flow, our objective is to generate a high-quality next-frame prediction with accurate pixel-level motion alignment. During the fine-tuning phase, the optical flow is provided directly by the dataset, whereas in the data synthesis phase, it is generated using the method described in Sec.~\ref{Sec.Method.flowgeneration}. 

Traditional conditioning-based image generation methods typically apply conditioning information directly to the latent embeddings. However, in our framework, the model requires both pixel-level appearance information from frame \( I_t \) and pixel-level motion information from flow \( F_{t \rightarrow t+1} \). Simple feature injection methods struggle to leverage both types of information simultaneously, often resulting in temporal inconsistency. To address this, we decompose the problem into two sub-challenges: (1) how to effectively utilize the pixel-level alignment information provided by the optical flow, and (2) how to efficiently incorporate the appearance information from the reference frame. As illustrated in Fig.~\ref{fig:conditioned-frame-generation}, \modelname leverages the reference appearance from \( I_t \) and motion from \( F_{t \rightarrow t+1} \) to synthesize the target frame \( I_{t+1} \). Our approach introduces two key novelties: (i) the use of \textit{pixel coordinate embeddings} to enhance pixel-optical alignment, and (ii) the incorporation of \textit{cross-frame information fusion} to improve generation quality.

\paragraph{Optical-Flow-Aware Coordinate Embedding}
\label{Sec.Method.Optical-Flow-AwareCoordinateEmbedding}
Unlike other conditioning methods~\cite{rombach2022highresolution,ho2020ddpm} that directly embed information into latent representations, optical flow describes the relative motion between frames rather than the content itself. Inspired by coordinate-based generation~\cite{seo2024genwarp,qiao2025genstereo}, we decompose the optical flow into dual coordinate embeddings: a \textit{canonical embedding} for the reference frame and a \textit{warped counterpart} for the target frame.

Specifically, we first construct a canonical 2D coordinate map \( C \in \mathbb{R}^{H \times W \times 2} \) normalized to \([-1, 1]\), which is transformed into Fourier features \( C_{t} = \mathbf{F}(C) \). Given the optical flow \( F_{t \rightarrow t+1} \), we generate the coordinate embeddings for the next frame by warping the canonical coordinates: \( C_{t+1} = \text{warp}(C_{t}, F_{t \rightarrow t+1}) \). These embeddings \((C_{t}, C_{t+1})\) are integrated into their respective frame features via convolutional layers, establishing strong pixel-level alignment and maintaining visual consistency.

\paragraph{Cross-Frame Information Conditioning}
To address the bottleneck of effective information fusion, we leverage a cross-view attention mechanism. We construct a cross-view attention module where the attention map captures the similarity between the reference frame and the frame being generated. Guided by the coordinate embeddings from Sec.~\ref{Sec.Method.Optical-Flow-AwareCoordinateEmbedding}, this mechanism learns pixel-level motion correspondences to maintain temporal and spatial consistency. 

We concatenate reference and target features within the attention mechanism. Specifically, the queries (\(q\)), keys (\(k\)), and values (\(v\)) are formulated as: 
\begin{equation}
    q = A_{t+1}, \quad k = [A_{t}, A_{t+1}], \quad v = [A_{t}, A_{t+1}],
\end{equation}
where \( A_{t} \) is derived from the reference U-Net conditioned on \((I_{t}, C_{t})\), and \( A_{t+1} \) is obtained from the denoising U-Net conditioned on \((I_{t+1}, C_{t+1})\). 

To seamlessly integrate generated and warped content, an Adaptive Fusion Module combines $I_{\text{gen}}$ and $I_{\text{warp}}$ based on local context:
\begin{equation}
    W = \sigma(f_\theta(\text{concat}(I_{\text{gen}}, I_{\text{warp}}, M))),
    \label{eq:fusion_weight}
\end{equation}
where $f_\theta$ is a convolution layer and $\sigma$ is the sigmoid activation, ensuring $W \in [0,1]$. This fusion module emphasizes warped content in high-confidence regions while relying on generated content in uncertain or occluded areas.

\subsection{Artificial Optical Flow Generation}
\label{Sec.Method.flowgeneration}

As illustrated in Fig.~\ref{fig:conditioned-frame-generation}, our artificial flow generator is built upon a 3D novel view synthesis (NVS) pipeline, which enables the creation of geometrically consistent motion fields from a single static image $I_t$.

\paragraph{3D Scene Reconstruction}
Given an input frame $I_t$, we first estimate its dense depth map using a pretrained monocular depth network $\mathcal{D}$:
\begin{equation}
D_t = \mathcal{D}(I_t).
\end{equation}
The depth model is fixed and never adapted to the target dataset. Using a pinhole camera model with a fixed field-of-view, each pixel $p_t = (u,v)$ with depth $d = D_t(u,v)$ is back-projected into 3D camera coordinates:
\begin{equation}
X_t = \text{BackProject}(p_t, d, P_t^{-1}),
\end{equation}
where $P_t$ denotes the source camera projection matrix.

\paragraph{Virtual Camera Motion}
To simulate motion, we define a virtual target camera $C_{t+1}$ by applying a randomly sampled rigid translation along the horizontal axis while keeping intrinsic parameters fixed. Let $V_t$ and $V_{t+1}$ denote the source and target view matrices, respectively. The relative camera transformation is:
\begin{equation}    
V_{\text{rel}} = V_{t+1} V_t^{-1}.
\end{equation}
Each 3D point is transformed to the target coordinate system:
\begin{equation}    
X_{t+1} = V_{\text{rel}} X_t.
\end{equation}

This virtual camera transformation (VCT) plays a critical role in reducing the domain gap between synthetic supervision and real-world benchmarks. By inducing motion through explicit 3D geometry rather than heuristic image warping, VCT generates geometrically consistent ground-truth flow while preserving the original visual distribution of the target dataset. Since improper motion scaling may introduce distributional shifts, we carefully align the virtual camera parameters with dataset-specific motion priors to ensure realistic displacement magnitudes.

\paragraph{Novel View Rendering and Flow Computation}
The transformed points are projected onto the target image plane using the same projection model:
\begin{equation}    
p_{t+1} = \text{Project}(X_{t+1}, P_{t+1}).
\end{equation}
Instead of performing naive forward splatting, we leverage a differentiable NVS warping module to compute a dense forward correspondence map $\mathcal{C}_{\text{forward}}$, which establishes pixel-aligned mappings between the synthesized frame $\tilde{I}_{t+1}$ and the source image $I_t$.

The artificial optical flow $\tilde{F}_{t\rightarrow t+1}$ is derived directly from this correspondence field. Specifically, for each target pixel location $p_{t+1}$, the flow is computed as the displacement between the target grid coordinate and its corresponding source coordinate:
\begin{equation}    
\tilde{F}_{t\rightarrow t+1}(p_{t+1}) 
= p_{t+1} - \mathcal{C}_{\text{forward}}(p_{t+1}).
\end{equation}

This procedure produces geometrically consistent and pixel-aligned triplets
\begin{equation}    
(I_t, \tilde{F}_{t\rightarrow t+1}, \tilde{I}_{t+1}),
\end{equation}
where $\tilde{I}_{t+1}$ is the synthesized next frame generated purely from $I_t$ and the sampled virtual camera motion.

\subsection{Inconsistent Pixel Filtering}
To mitigate the impact of misaligned pixels, we propose an \textit{inconsistent pixel filtering} strategy. Given a synthetic triplet $(I_{t}, \tilde{F}_{t \rightarrow t+1}, \tilde{I}_{t+1})$, we filter out unreliable regions to obtain a valid mask. Since motion is typically bounded, pixel displacements exceeding a threshold $Z$ or showing high photometric discrepancy are regarded as unreliable. We first estimate a warped frame $I'_{t+1}$ from $I_{t}$ and $\tilde{F}_{t \rightarrow t+1}$. The binary mask is computed as:
\begin{equation}
    \mathbf{M} = \mathbbm{1} \left( | \tilde{I}_{t+1} - I'_{t+1} | \leq Z \right).
\end{equation}
During fine-tuning, the reconstruction loss is computed only on valid regions:
\begin{equation}
    \mathcal{L} = \left\| ( \tilde{I}_{t+1} - I_{t+1} ) \odot \mathbf{M} \right\|_1,
\end{equation}
which effectively suppresses the influence of inconsistent pixels and improves robustness.
\section{Experiment}
\label{Sec.Experiment}

\subsection{Experiment Setup}
\paragraph{Datasets}
For the conditioned next-frame generation training phase, we fine-tune our model on the VKITTI2~\cite{cabon2020vkitti2} and TartanAir~\cite{tartanair2020iros} datasets, which provide large-scale triplets $(I_t, I_{t+1}, F_{t\rightarrow t+1})$ for supervised motion-consistent generation learning.

To construct the synthetic training data used for downstream refinement, we exclusively sample RGB frames from the \textbf{training splits only} of the target datasets. No ground-truth optical flow annotations from KITTI or Sintel are accessed at any stage of synthetic data construction or model optimization. The official validation/test splits and benchmark servers are strictly reserved for final evaluation and remain completely unseen during training. Detailed implementation procedures for synthetic data generation are described in Sec.~\ref{Sec.exp.imple}.

We evaluate the effectiveness of our framework on KITTI2012~\cite{Geiger2012CVPR}, KITTI2015~\cite{Menze2015CVPR}, and Sintel~\cite{ButlerECCV2012} using their standard evaluation protocols.

\paragraph{Foundamental Pretrained Models}
Due to computational resource limitations, we choose to fine-tune the pre-trained Stable Diffusion V1.5 UNet Model~\cite{rombach2022high} released by Stability AI on Hugging Face for the aforementioned datasets, aiming to generate the next frame conditioned on the previous one. In addition, for random manual optical flow generation, which requires depth estimation, we employ several state-of-the-art pretrained models, including Depth Anything~\cite{yang2024depth}, Depth Anything V2~\cite{depth_anything_v2} and ZoeDepth~\cite{bhat2023zoedepth}.

\paragraph{Implementation Detail}
\label{Sec.exp.imple}
All experiments are conducted using the PyTorch framework~\cite{paszke2019pytorch} on a Slurm-based computing cluster equipped with mixed NVIDIA GPU resources, including A100, L40S, and H100.

We first fine-tune the pretrained Stable Diffusion model for three epochs on a mixture of VKITTI2 and TartanAir datasets, applying a fivefold higher sampling ratio to VKITTI2 to improve dataset balance. During this stage, \modelname learns to establish conditional next-frame generation behavior that captures optical flow–consistent motion patterns between consecutive frames.

To construct the synthetic training set for downstream optical flow refinement, we exclusively sample single RGB frames $I_t$ from the \textbf{training split only} of the target dataset (i.e., Sintel or KITTI). No consecutive ground-truth frame pairs or optical flow annotations are accessed. The official validation/test splits and benchmark servers are strictly reserved for final evaluation and remain completely unseen during training.

Following the geometric flow generation strategy introduced in Sec.~\ref{Sec.Method.flowgeneration}, we first construct artificial optical flow fields $F_{t\rightarrow t+1}$ using randomly sampled geometric transformations. A fixed pretrained depth estimation model is employed solely to provide structural priors for generating geometrically consistent artificial flow; it is not trained or adapted on the target dataset.

Given the sampled frame $I_t$ and the generated artificial flow $F_{t\rightarrow t+1}$, the trained conditioned next-frame generation model synthesizes the subsequent frame $I_{t+1}$. This results in aligned triplets $(I_t, I_{t+1}, F_{t\rightarrow t+1})$, where $F_{t\rightarrow t+1}$ denotes the artificial flow used to drive the generation process.

Repeating this procedure $N=5{,}000$ times yields a synthetic dataset
\begin{equation}    
D = \{(I_t^{(i)}, I_{t+1}^{(i)}, F_{t\rightarrow t+1}^{(i)})\}_{i=1}^{N},
\end{equation}
which serves as pseudo-supervision without relying on any ground-truth optical flow labels.

Finally, using the synthetic dataset $D$, we fine-tune baseline models that were originally trained under unsupervised objectives for one additional epoch in this pseudo-supervised setting. Performance is then evaluated separately on the official KITTI and Sintel benchmarks using their respective evaluation protocols, ensuring strict separation between training data and evaluation data.
\subsection{Results}

\paragraph{Next Frame Generation Quality}

\begin{table*}[htbp]
\centering
\resizebox{\textwidth}{!}{%
\begin{tabular}{l|c|ccc|ccc} 
\toprule
\multirow{2}{*}{\textbf{Methods}} & {\textbf{Pretrained}} & \multicolumn{3}{c|}{\textbf{Middlebury 2014}} & \multicolumn{3}{c}{\textbf{KITTI 2015}} \\ 
\cmidrule{3-8} 
& {\textbf{Model}}& \textbf{PSNR} $\uparrow$ & \textbf{SSIM} $\uparrow$ & \textbf{LPIPS} $\downarrow$ & \textbf{PSNR} $\uparrow$ & \textbf{SSIM} $\uparrow$ & \textbf{LPIPS} $\downarrow$ \\
\midrule
Leave blank &-&11.328&0.315&0.450&12.980&0.374&0.313\\
Stretch &-&14.842&0.432&0.285&14.757&0.429&0.212\\
3D Photography~\cite{Shih3DP20}&-&14.190&0.427&0.275& 14.540&0.398&0.210\\
RePaint~\cite{lugmayr2022repaintinpaintingusingdenoising}&-&15.102&0.462&0.311&15.056&0.462&0.251\\
SD-Inpainting~\cite{stacchio2023stableinpainting}&-&15.740&0.412&0.311&9.792&0.230&0.652\\
StereoDiffusion~\cite{Wang_2024}&-&15.456&0.468&0.231&15.679&0.481&0.205\\
\midrule
Ours-\modelname&SD1.5&N/A&N/A&N/A&17.854&0.552&0.268\\
\midrule
Ours-\modelname&SD1.5+RAFT~\cite{teed2020raft}&19.243&0.653&0.217&N/A&N/A&N/A\\
Ours-\modelname&SD1.5+WAFT~\cite{wang2025waftwarpingalonefieldtransforms}&20.864&0.700&0.168&N/A&N/A&N/A\\
\bottomrule
\end{tabular}}
\caption{Quantitative comparison of motion-conditioned image generation methods on the Middlebury 2014 and KITTI 2015 datasets. Evaluation metrics include PSNR ($\uparrow$), SSIM ($\uparrow$), and LPIPS ($\downarrow$). For the Middlebury 2014 dataset, we report results using optical flow estimated by RAFT and WAFT, while the KITTI 2015 dataset provides ground-truth optical flow for evaluation. Qualitative results can be found in Supp~\ref{sec:appendix_implementation}}
\label{tab:quantitaive_results_conditioning_image_generation}
\end{table*}

The proposed high-quality next-frame generation framework serves as the foundation of our approach. To evaluate its out-of-domain generalization capability, we conduct quantitative experiments on the KITTI 2015 and Middlebury 2014 datasets. The KITTI 2015 dataset provides ground-truth next frames and optical flow, while the Middlebury 2014 dataset does not. For the latter, we utilize two representative supervised optical flow estimators, RAFT~\cite{teed2020raft} and WAFT~\cite{wang2025waftwarpingalonefieldtransforms}, to generate pseudo ground-truth optical flow between stereo pairs. Both models are pretrained on the KITTI datasets and yield reasonable optical flow estimates on Middlebury 2014. Evaluation is performed using three standard metrics: PSNR, SSIM, and LPIPS~\cite{zhang2018unreasonable}. PSNR measures pixel-level reconstruction fidelity, SSIM assesses structural and photometric consistency, and LPIPS evaluates perceptual similarity in the feature space.

As no existing work directly targets next-frame generation, we include several stereo generation methods for comparison, as they pursue a related goal of synthesizing novel views under relative motion. Furthermore, since the optical flow annotations in KITTI 2015 are relatively sparse, while our subsequent synthetic datasets provide fully dense, artificially generated flow, we adopt a random optical flow dropout strategy inspired by GenStereo~\cite{qiao2025genstereo}. Specifically, 10\% of the flow points are randomly masked during training to enhance robustness and better match the sparsity characteristics of KITTI 2015. The detailed implementation of this strategy is provided in the appendix. For the Middlebury 2014 dataset, this issue does not arise, as we employ a pretrained dense optical flow estimator to obtain the flow conditioning.

As shown in Tab.~\ref{tab:quantitaive_results_conditioning_image_generation}, compared to state-of-the-art stereo generation methods, our \modelname tackles a more challenging next-frame generation task while still achieving competitive performance. Specifically, \modelname attains PSNR, SSIM, and LPIPS scores of 17.854, 0.552, and 0.268, respectively. Furthermore, when using optical flow predicted by pretrained models, \modelname achieves improved results of 20.864, 0.700, and 0.168 for PSNR, SSIM, and LPIPS, respectively. Notably, during the artificial optical flow generation phase, \modelname produces dense optical flow, indicating that the actual next-frame generation quality may surpass the reported metrics. These results demonstrate the effectiveness of our approach, highlighting its accurate pixel–flow alignment and validating the use of synthetic datasets for robust model evaluation. More qualitative results can be found in Supp~\ref{sec:appendix_implementation}.

\paragraph{Zero-shot Optical Flow Estimation Enhancement}
\begin{table*}[t]
    \centering
    \begin{tabular}{lcccc}
    \toprule
    \multirow{2}{*}{\textbf{Method}} 
    & \multicolumn{2}{c}{\textbf{Without \modelname}} 
    & \multicolumn{2}{c}{\textbf{With \modelname}} \\
    \cmidrule(lr){2-3} \cmidrule(lr){4-5}
    & EPE$\downarrow$ & Fl-all$\downarrow$ & EPE$\downarrow$ & Fl-all$\downarrow$ \\
    \midrule
    RAFT-chairs~\cite{teed2020raft} & 9.82 & 37.55 & 8.18 & 20.59 \\
    GMFlowNet-things~\cite{zhao2022globalmatchingoverlappingattention} & 4.43 & 16.70 &4.24&16.68\\
    FlowFormer-chairs~\cite{huang2022flowformer} & 10.14 & 34.46 & 7.45&21.19\\
    Flowformer++-chairs~\cite{shi2023flowformer++} & 10.45 & 36.97 &7.98&22.01\\
    FlowDiffuser-things~\cite{luo2024flowdiffuser} & 3.61 & 11.8 &3.52&10.69\\
    WAFT-chairs~\cite{wang2025waftwarpingalonefieldtransforms} & 36.73 & 85.30 &33.41&82.60\\
    \bottomrule
    \end{tabular}
\caption{Quantitative zero-shot evaluation on the KITTI 2015 benchmark. We report the EPE and Fl-all metrics for six representative optical flow models, both with and without integrating our proposed \modelname. Across all methods, incorporating \modelname consistently reduces error, demonstrating its strong generalization ability and compatibility with diverse architectures. Lower values indicate better performance.}
\label{tab:exp-zero-shot-wide-comparison}
\end{table*}
As described in Sec.~\ref{Sec.Method.flowgeneration}, we leverage our next-frame generation model and artificial optical flow generation method to synthesize data samples using real frames randomly sampled from the target datasets. Each synthesized sample contains dense optical flow and pixel-level well-aligned next frames. Specifically, we generated $N=5000$ such triplets for the KITTI2012, KITTI2015, and Sintel datasets. Following prior work, we train the baseline models without modifying their architectures. We then compare models trained on our synthetic datasets with baseline models that were not trained on the target datasets in a supervised manner. Additionally, for certain supervised methods, we can also compare against models that provide publicly available checkpoints, which were trained on datasets other than the target datasets.

As shown in Tab.~\ref{tab:exp-zero-shot-wide-comparison}, we report comprehensive comparisons against previous state-of-the-art models under zero-shot settings on the Sintel and KITTI2015 datasets, which can be viewed as an unsupervised evaluation scenario. All baseline results are obtained from the official open-source checkpoints provided by the respective authors. For KITTI2015, we evaluate model performance using both EPE and the Fl-all metric. As the results indicate, \modelname achieves consistent improvements across six different frameworks, reducing EPE by 1.49 and Fl-all by 7.00, respectively.

\paragraph{Synthetic Dataset Generation Overhead}\mbox{}\\
\textbf{Data Generation:} The synthetic dataset construction is performed as a one-time offline preprocessing step. For our experiments, generating $N=5{,}000$ triplets requires less than 4 hours on a node with 4 GPUs and is fully parallelizable across devices. Importantly, this cost is incurred once and can be amortized across multiple downstream training runs.

In contrast, acquiring real-world optical flow supervision necessitates complex multi-sensor setups (e.g., synchronized stereo rigs or LiDAR systems), precise calibration, controlled motion capture, and extensive post-processing. Such pipelines are logistically demanding, time-consuming, and financially expensive, especially for large-scale or diverse scene coverage.

Our approach replaces physical data acquisition with purely computational synthesis, eliminating hardware dependencies and manual collection efforts. Moreover, the downstream refinement stage requires only a single additional training epoch, keeping the overall computational overhead modest relative to conventional supervised data collection and annotation workflows.

\textbf{Finetuning:} \modelname{} adopts standard optical flow estimation models without architectural modifications. Therefore, finetuning latency, FLOPs, and parameter counts remain identical to the underlying unsupervised baseline models.

\subsection{Ablation Study}
\paragraph{Coordinate embedding and cross-view attention}
\begin{wraptable}{r}{0.55\textwidth} 
    \centering
    \caption{\textbf{Ablation Study.}}
    \label{tab:different_components}
    \resizebox{0.55\textwidth}{!}{
        \begin{tabular}{l|ccc|ccc}
            \toprule
            & \multicolumn{3}{c|}{\textbf{Middlebury}} & \multicolumn{3}{c}{\textbf{KITTI 2015}} \\ \cline{2-7} 
            & \textbf{PSNR} & \textbf{SSIM} & \textbf{LPIPS} & \textbf{PSNR} & \textbf{SSIM} & \textbf{LPIPS} \\ \midrule
            w/o CVA & 17.93 & .417 & .314 & 15.36 & .291 & .290 \\
            w/o Coord & 17.15 & .508 & .291 & 14.63 & .384 & .314 \\ \midrule
            \textbf{Ours} & \textbf{19.24} & \textbf{.653} & \textbf{.217} & \textbf{17.85} & \textbf{.552} & \textbf{.268} \\ \bottomrule
        \end{tabular}
    }
\end{wraptable}
We validate the necessity of both components in Table~\ref{tab:different_components}. Removing \textbf{Coordinate Embeddings} degrades geometric alignment, leading to significant drops in PSNR. Disabling \textbf{Cross-view Attention} breaks multi-view consistency, causing the model to process frames independently and resulting in structural flickering. The full model combines these to ensure both geometric accuracy and temporal consistency.

\paragraph{Effect of Depth Estimation Model}
Addressing concerns regarding pre-trained model bias, Table~\ref{tab:different_MDE} shows consistent performance improvements when upgrading from ZoeDepth to Depth Anything V2. This indicates that our framework does not impose a fixed performance ceiling but instead benefits directly from advances in monocular depth estimation. While errors from pre-trained depth models may propagate to flow training, our results suggest that such effects can be effectively mitigated by adopting stronger depth estimators.
\begin{table}[htbp]
    \centering
    \begin{minipage}[t]{0.40\textwidth}
        \centering
        \resizebox{\linewidth}{!}{
            \begin{tabular}{l c c} 
                \toprule
                \textbf{Use Filtering} & \textbf{KITTI EPE} & \textbf{KITTI Fl-all}\\
                \midrule
                W/O &9.29&27.64\\
                W/ Threshold=50&8.51&24.83\\
                W/ Threshold=30&8.18&20.59\\
                W/ Threshold=20&8.41&25.56\\
                \bottomrule
            \end{tabular}
        }
        \caption{Impact of the inconsistent pixel filtering threshold.}
        \label{tab:exp-ablation-filtering-trick}
    \end{minipage}
    \hfill
    \begin{minipage}[t]{0.55\textwidth}
        \centering
        \resizebox{\linewidth}{!}{
            \begin{tabular}{c|ccc|ccc}
                \toprule
                \textbf{Pretrained} & \multicolumn{3}{c|}{\textbf{Middlebury 2014}} & \multicolumn{3}{c}{\textbf{KITTI 2015}} \\ \cline{2-7} 
                \textbf{MDE Model} & \textbf{PSNR} $\uparrow$ & \textbf{SSIM} $\uparrow$ & \textbf{LPIPS} $\downarrow$ & \textbf{PSNR} $\uparrow$ & \textbf{SSIM} $\uparrow$ & \textbf{LPIPS} $\downarrow$ \\ \midrule
                ZoeDepth & 17.583 & 0.627 & 0.233 & 16.396 & 0.527 & 0.292 \\
                Depth Anything V1 & \textbf{19.245} & 0.648 & 0.220 & 17.852 & 0.549 & 0.274 \\ \midrule
                Depth Anything V2 & 19.243 & \textbf{0.653} & \textbf{0.217} & \textbf{17.854} & \textbf{0.552} & \textbf{0.268} \\ \bottomrule
            \end{tabular}
        }
        \caption{{Ablation of Depth Estimation Backbones.}}
        \label{tab:different_MDE}
    \end{minipage}
\end{table}

\paragraph{Inconsistent Pixel Filtering}

Table~\ref{tab:exp-ablation-filtering-trick} reports the impact of applying inconsistent pixel filtering during synthetic data generation. This strategy aims to remove unreliable or highly inconsistent pixels, which are typically caused by inaccurate optical flow estimation, to improve the quality of the generated training pairs.

Compared with the baseline without filtering, applying a moderate filtering threshold significantly improves performance, reducing both EPE and Fl-all. In particular, using a threshold of 30 yields the best results, achieving a substantial improvement of 1.11 in EPE and 7.05 in Fl-all. This demonstrates that filtering out severely inconsistent pixels helps the model avoid learning from corrupted motion cues, leading to more stable fine-tuning.

Interestingly, overly strict filtering (e.g., threshold = 20) degrades performance, likely because excessive pixel removal reduces the diversity and completeness of the synthetic training samples. These results highlight the importance of balancing data cleanliness and diversity: moderate filtering enhances optical-flow consistency, whereas aggressive filtering harms the model by removing too much informative content.

\section{Conclusion}
In this work, we introduced \textbf{\modelname}, a novel framework for synthesizing large-scale, pixel-aligned frame--flow pairs to enable supervised optical flow training from unlabeled videos. By leveraging depth-guided pseudo optical flow and a next-frame generation model, our approach produces high-fidelity synthetic data that captures accurate motion correspondences. To further enhance training reliability, we proposed an inconsistent pixel filtering strategy to remove unreliable pixels, thereby improving fine-tuning performance on downstream tasks. Extensive experiments on KITTI2012, KITTI2015, and Sintel datasets demonstrate that {\modelname} significantly narrows the performance gap between unsupervised and fully supervised optical flow methods. Our framework provides a scalable, annotation-free solution for optical flow estimation, reducing the dependency on expensive ground-truth labels while maintaining high accuracy.

\paragraph{Limitations and Future Work}
Our approach still faces several limitations. The quality of the synthesized supervision depends on the reliability of the pseudo optical flow and next-frame generation model, which can degrade under large motion or heavy occlusions. Inconsistent pixel filtering also removes some informative regions, potentially reducing data diversity.

Future work includes improving robustness in challenging motion regimes, enforcing temporal consistency over longer sequences, and extending \modelname{} to broader video understanding tasks such as action recognition and frame interpolation.

\bibliographystyle{plain} 
\bibliography{main}

\newpage
\appendix
\clearpage

\section{Supplementary Material}
\label{sec:appendix_implementation}
\subsection{Conditioned Frame Generation}
\label{subsec:training_details}
\paragraph{Architecture and Initialization.} 
Our framework is built upon the pre-trained Stable Diffusion v2.1 text-to-image model. We utilize the \texttt{sd-vae-ft-mse} model as our Variational Autoencoder (VAE) and the image encoder from \texttt{sd-image} \texttt{-variations-diffusers}. During the training phase, both the VAE and the CLIP image encoder parameters are frozen. We employ two UNet architectures: a Reference UNet and a Denoising UNet. The Reference UNet is initialized from the Stable Diffusion checkpoint; to preserve pre-trained semantic knowledge, we freeze the parameters in its highest-level upsampling block (\texttt{up\_blocks.3}) while fine-tuning the rest. The Denoising UNet is fully trainable. The Pose Guider and the Adaptive Fusion Layer are initialized from scratch.

\paragraph{Training Data and Preprocessing.} 
We train our model primarily on the VKITTI2 and TartanAir datasets. Input images are resized to a resolution of $512 \times 512$. To improve model robustness, we implement data augmentation strategies, including random cropping and resizing during the data loading process. Ground optical flow data from the dataset is used to generate the coordinate embeddings used for geometric conditioning.

\paragraph{Hyperparameters and Optimization.} 
The training is conducted on NVIDIA A100 GPUs. We use the AdamW optimizer with $\beta_1 = 0.9$, $\beta_2 = 0.999$, a weight decay of $1.0 \times 10^{-2}$, and $\epsilon = 1.0 \times 10^{-8}$. The learning rate is fixed at $1.0 \times 10^{-5}$ following a constant schedule with a warmup of 1 step. The global batch size is set to 2 per GPU, and the model is trained for a total of 1 epoch. To reduce memory footprint and accelerate training, we utilize FP16 mixed-precision training.

\paragraph{Noise Scheduler and Objectives.} 
We employ a DDIM noise scheduler with a \texttt{scaled\_linear} beta schedule, where $\beta_{start}=0.00085$ and $\beta_{end}=0.012$. The model is trained using the velocity prediction ($v$-prediction) objective, coupled with the Zero-SNR strategy to enhance stability. Our loss function consists of two components: a diffusion loss and a pixel-wise reconstruction loss. The diffusion loss is computed using Mean Squared Error (MSE) with Min-SNR weighting ($\gamma = 5.0$) to balance the loss magnitude across different timesteps. Additionally, we enable a pixel-level MSE loss (\texttt{pixel\_loss}) between the decoded predicted image and the ground truth target to further enforce visual fidelity.

\paragraph{Conditioning and Guidance.} 
To facilitate classifier-free guidance during inference, we randomly drop the CLIP image embeddings with a probability of $p=0.1$ during training. Geometric conditions, including disparity-warped coordinates and images, are encoded by the Pose Guider and injected into the Denoising UNet.

\subsection{Artificial Optical Flow Generation}
\label{subsec:optical_flow_generation}

\paragraph{Data Preprocessing.}
Our generation pipeline is built upon the KITTI 2012 and Sintel dataset. In the initial stage, raw images are extracted from the training set and organized into a unified directory structure to serve as the input source for the subsequent generation process.

\paragraph{Depth Estimation.}
To recover geometric information from monocular images, we utilize the \textbf{Depth Anything V2} model. Specifically, we employ the large Vision Transformer variant (\texttt{vitl}) as the encoder, loaded with weights fine-tuned for outdoor scenes (\texttt{metric\_vkitti}). Considering the characteristics of outdoor driving scenarios, the maximum depth range is set to 80 meters.

\paragraph{Novel View Synthesis Configuration.}
We employ the \textbf{GenWarp} framework for novel view synthesis, utilizing the \texttt{multi1} checkpoint. To ensure numerical stability and prevent precision-related artifacts during the warping process, we explicitly disable half-precision weights, enforcing the model to operate in full precision (Float32). Input images are center-cropped to the shorter side and resized to a resolution of $512 \times 512$ before processing.

\paragraph{Camera Motion Simulation.}
To generate image pairs for optical flow computation, we simulate lateral camera motion. The source camera is positioned at the origin of the world coordinate system. The target camera pose is determined by applying a random translation along the X-axis. The translation distance $d$ is sampled from a uniform distribution $U(0.8, 1.2)$, with the direction (left or right) chosen randomly. The vertical fields of view (FOVY) for the projection matrix are fixed at $29.2^{\circ}$ and $26.5^{\circ}$ for KITTI and Sintel, respectively.

\paragraph{Optical Flow Calculation.}
The artificial optical flow is derived directly from geometric projection rather than traditional matching algorithms. The GenWarp model outputs a correspondence grid representing the mapping between the source and target views in normalized coordinates $[-1, 1]$. We compute the normalized optical flow displacement by calculating the difference between this predicted correspondence grid and the regular identity grid of the target view. The resulting flow data is saved in \texttt{.npy} format for downstream training or evaluation.

\section{Additional Visualization Results}

\begin{figure*}[h]
    \centering
    \includegraphics[width=\textwidth]{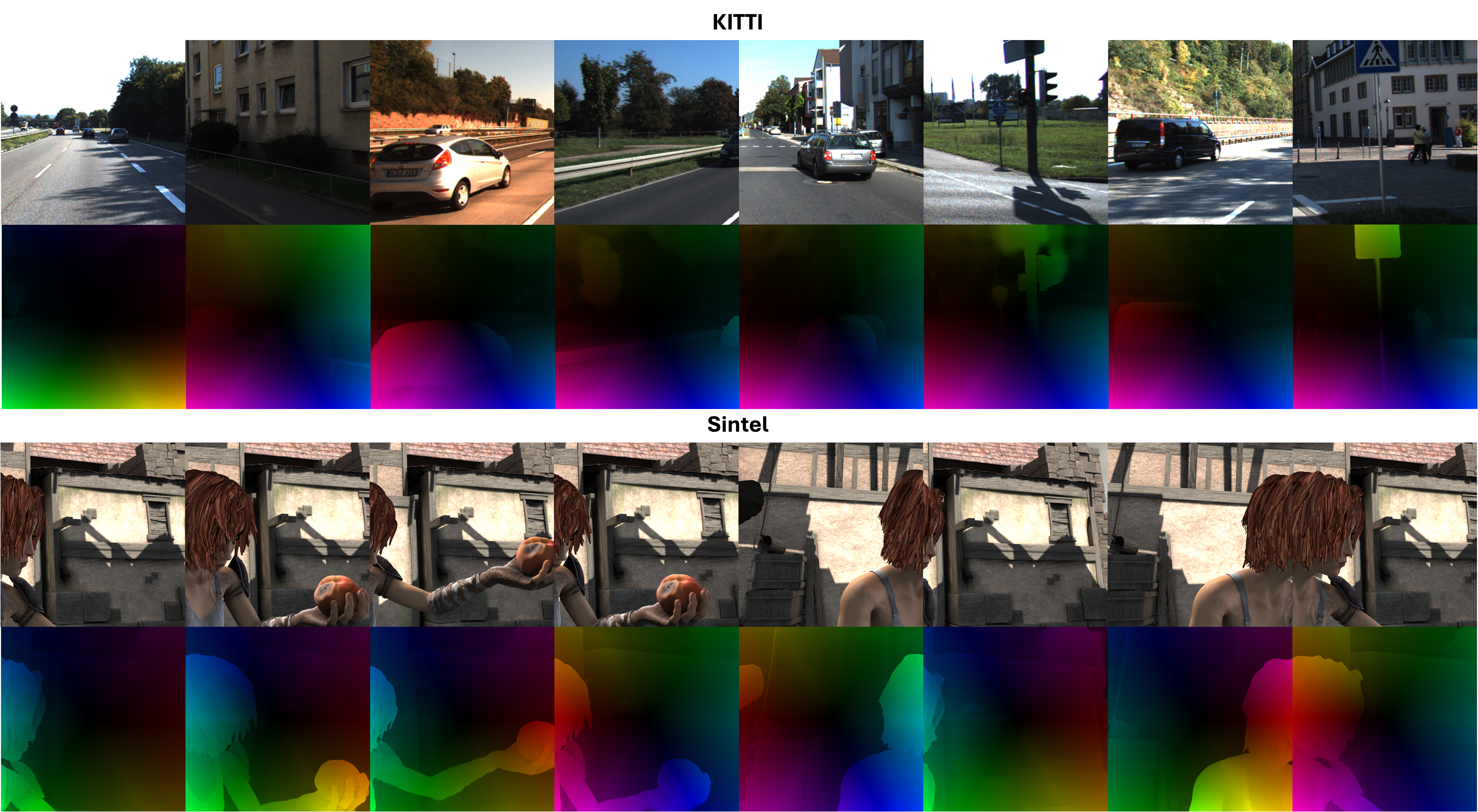}
    \caption{\textbf{Qualitative results of optical flow generation on the KITTI (top) and Sintel (bottom) datasets.} In each panel, the first row displays the conditioning input frame, while the second row visualizes a corresponding optical flow map randomly sampled from our model. The results highlight the model's ability to generate dense, structurally aligned flow predictions across varying scene complexities.}
    \label{fig:sup-optical-flow}
\end{figure*}

\begin{figure*}[h]
    \centering
    \includegraphics[width=\textwidth]{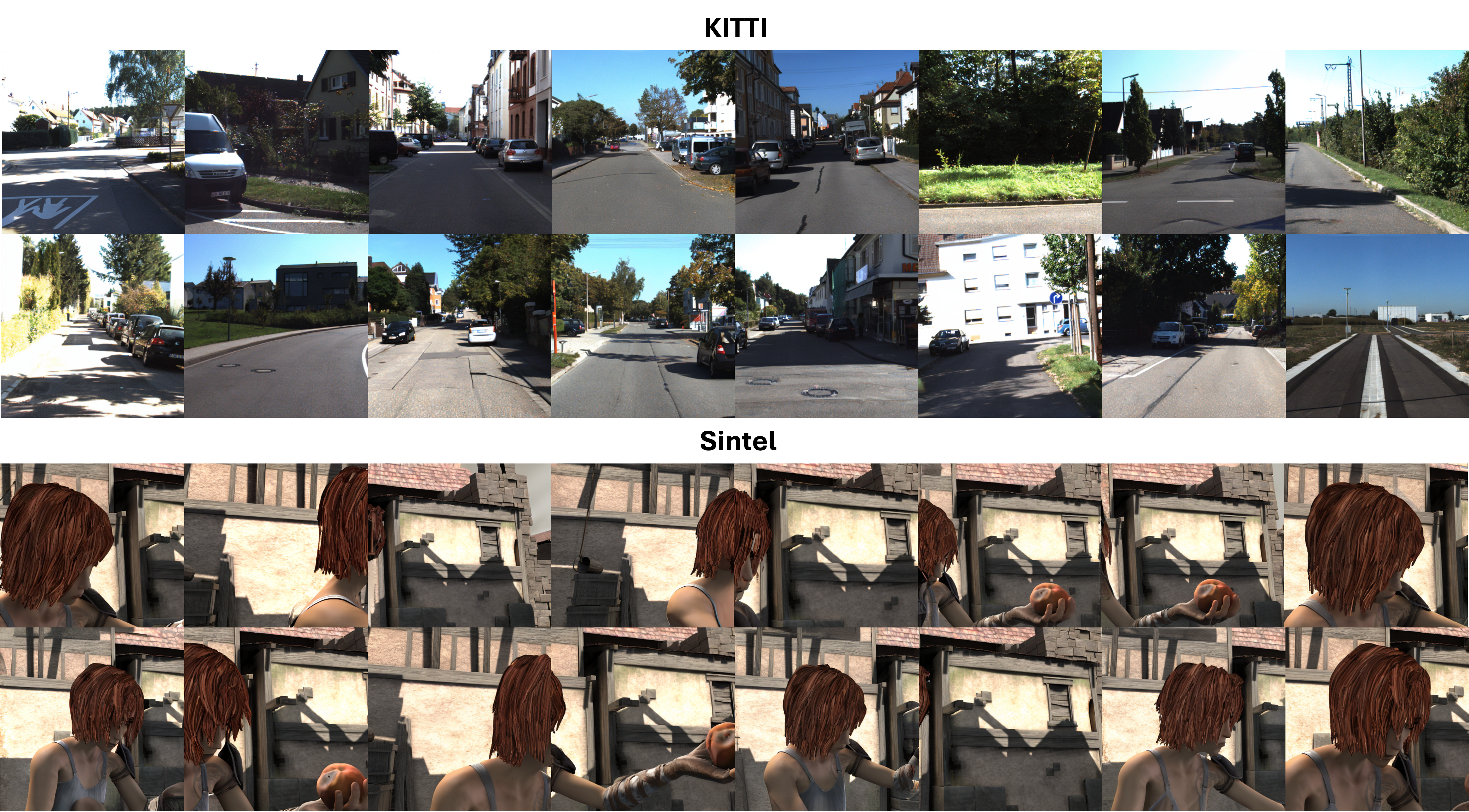}
    \caption{\textbf{Visualization of next-frame generation results.} The figure demonstrates the temporal consistency of our method on the KITTI and Sintel benchmarks. The top rows show the reference input frames, while the bottom rows display the synthesized next frames. Our framework effectively preserves texture details and object geometry during the generation process.}
    \label{fig:sup-next-frame}
\end{figure*}

In this section, we provide additional qualitative results in Fig.~\ref{fig:sup-next-frame} and Fig.~\ref{fig:sup-optical-flow} to demonstrate the efficacy and generalization capability of our proposed framework. We visualize the model's performance across two distinct domains: real-world driving scenes (KITTI) and synthetic animated sequences (Sintel). These examples illustrate the model's ability to synthesize high-fidelity optical flow maps and maintain structural consistency in next-frame generation tasks.

%
%
%
%

\end{document}